\pdfoutput=1

\documentclass[11pt]{article}

\usepackage[final]{acl}
\usepackage{times}
\usepackage{latexsym}

\usepackage[T1]{fontenc}

\usepackage[utf8]{inputenc}
\usepackage{amsmath}
\usepackage{microtype}
\usepackage{booktabs}
\usepackage{multirow}
\usepackage{makecell}
\usepackage{algorithmic}
\usepackage{colortbl}  
\usepackage{xcolor}
\usepackage{algorithm}
\usepackage{inconsolata}

\usepackage{graphicx}

\title{Fast Quiet-STaR: Thinking Without Thought Tokens}

\author{
 \textbf{Wei Huang\textsuperscript{1}}\thanks{These authors contributed equally to this work.},
 \textbf{Yizhe Xiong\textsuperscript{2,3}}\footnotemark[1],
 \textbf{Xin Ye\textsuperscript{4}}\footnotemark[1],
 \textbf{Zhijie Deng\textsuperscript{5}},
\\
 \textbf{Hui Chen\textsuperscript{2,3}$^{\dagger}$},
 \textbf{Zijia Lin\textsuperscript{2}$^{\dagger}$},
 \textbf{Guiguang Ding\textsuperscript{2,3}}
\\
\\
 \textsuperscript{1}School of Computer Science, Beijing University of Posts and Telecommunications, \\
 \textsuperscript{2}Tsinghua University, \\
 \textsuperscript{3}Beijing National Research Center for Information Science and Technology (BNRist), \\
 \textsuperscript{4}Kuaishou Technology, 
 \textsuperscript{5}Shanghai Jiao Tong University, Shanghai, China
\\
 \small{
   \textbf{Correspondence$^{\dagger}$:} \href{mailto:jichenhui2012@gamil.com}{jichenhui2012@gamil.com}, \href{mailto:linzijia07@tsinghua.org.cn}{linzijia07@tsinghua.org.cn}
 }
}

\begin{document}
\maketitle
\begin{abstract}
Large Language Models (LLMs) have achieved impressive performance across a range of natural language processing tasks. However, recent advances demonstrate that further gains—particularly in complex reasoning tasks—require more than merely scaling up model sizes or training data. One promising direction is to enable models to ``think''  during the reasoning process. Recently, Quiet-STaR significantly improves reasoning by generating token-level thought traces, but incurs substantial inference overhead. In this work, we propose Fast Quiet-STaR, a more efficient reasoning framework that preserves the benefits of token-level reasoning while reducing computational cost. Our method introduces a curriculum-learning-based training strategy that gradually reduces the number of thought tokens, enabling the model to internalize more abstract and concise reasoning processes. We further extend this approach to the standard Next Token Prediction (NTP) setting through reinforcement learning-based fine-tuning, resulting in Fast Quiet-STaR NTP, which eliminates the need for explicit thought token generation during inference. Experiments on four benchmark datasets with Mistral 7B and Qwen2.5 7B demonstrate that Fast Quiet-STaR consistently outperforms Quiet-STaR in terms of average accuracy under the same inference time budget. Notably, Fast Quiet-STaR NTP achieves an average accuracy improvement of 9\% on Mistral 7B and 5.7\% on Qwen2.5 7B, while maintaining the same inference latency. 
\end{abstract}
\section{Introduction}
Large Language Models (LLMs) \cite{gpt4,llama3} have achieved remarkable progress in recent years by pretraining models with billions of parameters on massive datasets. However, merely scaling up model size or increasing the amount of training data is insufficient for enabling strong performance on tasks that require complex reasoning or long-term planning. To further enhance model capabilities, one promising direction is to enable models to engage in autonomous ``thinking'' before producing final answers. Recently, a growing body of research has explored this paradigm to strengthen the reasoning abilities of LLMs. Notably, models such as OpenAI o1 \cite{openaio1}, DeepSeek-R1 \cite{deepseekr1}, QwQ \cite{QWQ}, and Kimi-1.5 \cite{Kimi1.5} have demonstrated impressive performance across a variety of challenging tasks, such as mathematical competition problems \cite{math500,gsm8k}. 
\begin{figure}[t]
  \includegraphics[width=0.9\linewidth]{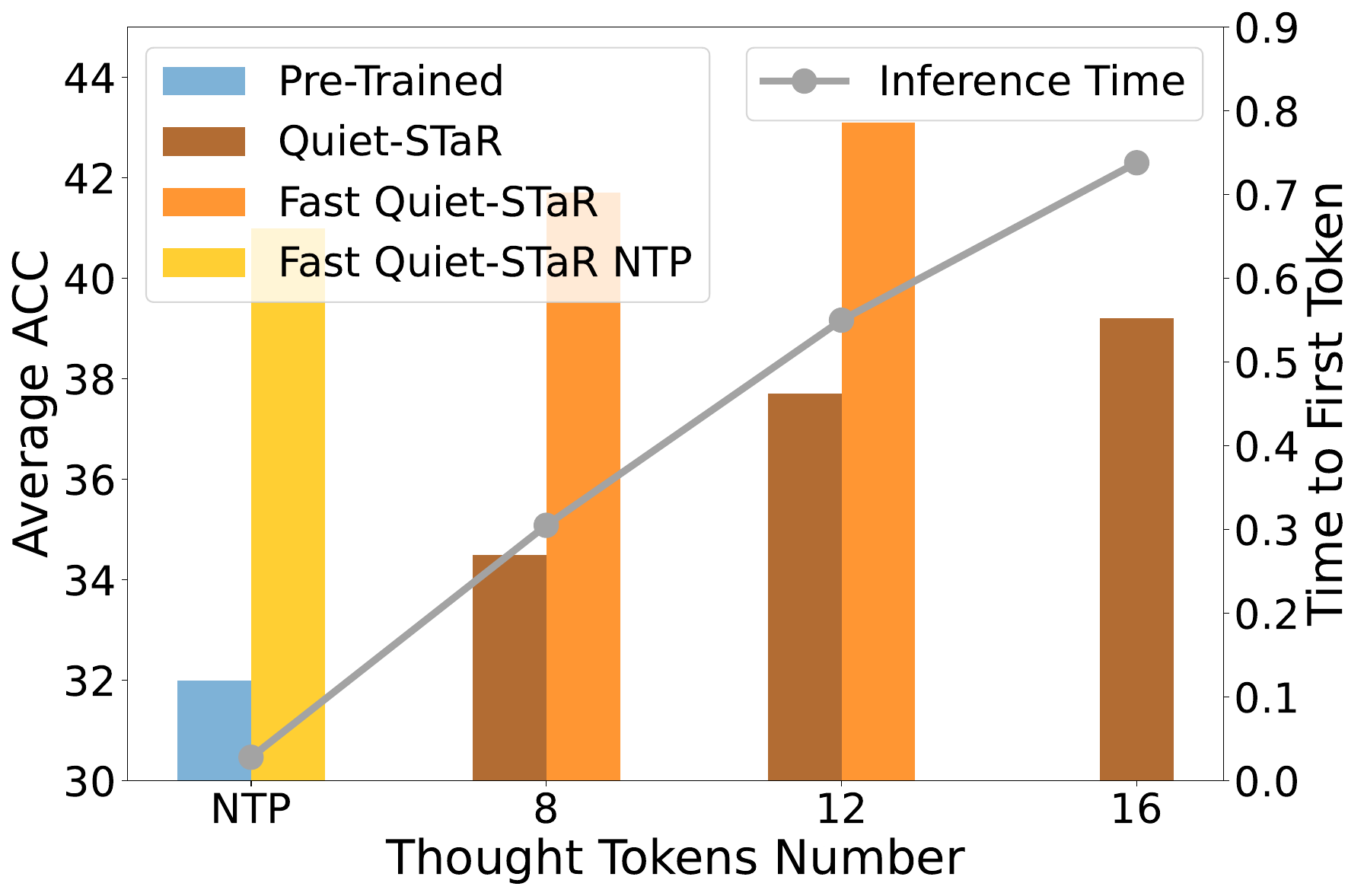}
  \caption{Performance comparison between Fast Quiet-STaR, Quiet-STaR and the pre-trained model (NTP). ``Inference Time'' represents the Time-to-First-Token (TTFT) of each model variant. Note that with the same number of thought tokens, Fast Quiet-STaR shares the same inference time with Quiet-STaR, but enjoys a significant performance boost. Additionally, Fast Quiet-STaR can be extended to the NTP setting, improving model performance without additional inference time overhead.}
  \label{Intro} 
\end{figure}

Recently, Quiet-STaR (Quiet Self-Taught Reasoner) \cite{quiet-star} has been proposed as a novel reasoning paradigm that shifts the thinking process from the problem level to a finer, token-level granularity. In Quiet-STaR, before predicting the next token, the model first generates a sequence of intermediate thought trace (represented as \textit{<|start\_of\_thought|>,thought\_token1, thought\_token2, ... <|end\_of\_thought|>}), based on which the model predicts the next token. Compared to other approaches, Quiet-STaR can significantly enhance the model's reasoning ability through a lightweight unsupervised training process. For instance, it achieves a 10\% performance gain on CommonsenseQA \cite{talmor2018commonsenseqa} for the Mistral 7B \cite{mistral7b} by continue pre-training with only \textbf{0.2M} tokens, demonstrating remarkable improvements through efficient training.

Although Quiet-STaR significantly enhances the model's reasoning capabilities, it substantially increases inference overhead due to the requirement of generating a thought trace for every token. As shown in Figure \ref{Intro}, even when using only 8 thought tokens, the average Time-To-First-Token (TTFT) of Quiet-STaR remains over \textbf{10 times higher} than that of conventional next token prediction (NTP) models. Despite the high inference costs, these thought tokens cannot be directly reduced or eliminated as they are the main contributor to performance improvements. For example, as shown in Figure \ref{Intro}, halving  the number of thought tokens from 16 to 8 leads to a 4.7\% accuracy drop. Such dilemma on efficiency severely undermines the practical value of Quiet-STaR. 

Prior work has shown that LLMs are capable of skipping reasoning steps by omitting non-essential steps without sacrificing overall reasoning performance \cite{liu2024can}. Inspired by this, we believe that within the Quiet-STaR reasoning paradigm, \textbf{the model can maintain its strong reasoning abilities obtained from long thought-trace training by compressing the number of thought tokens and keeping only a more abstract thought trace}. To improve the efficiency of the Quiet-STaR reasoning paradigm with minimal performance degradation, we propose \textbf{Fast Quiet-STaR}. We employ a multi-stage training strategy that progresses from easy to hard. That is, we gradually guide the model from generating a detailed thought trace using more thought tokens to generating a concise thought trace using fewer thought tokens. To further accelerate the Quiet-STaR inference paradigm to NLP-level efficiency, we employ a reinforcement learning-based fine-tuning strategy for Fast Quiet-STaR model under the NTP setting. The resulting Fast Quiet-STaR NTP model preserves the original thinking abilities of Fast Quiet-STaR while eliminating reliance on generating explicit thought trace during inference. 

We evaluate our method on two open-source models, Mistral 7B \cite{mistral7b} and Qwen2.5 7B \cite{qwen25} across four public datasets. Extensive experiments show that, given the same number of thought tokens(same inference time), Fast Quiet-STaR achieves substantial performance gains over Quiet-STaR. Furthermore, under equivalent inference time, Fast Quiet-STaR NTP improves the average accuracy by 9\% on Mistral 7B and 5.7\% on Qwen2.5 7B compared to the original pre-trained models.

We summarize our contribution as follows:
\begin{itemize}
    \item We propose Fast Quiet-STaR, a novel training paradigm that compresses token-level thought traces to significantly reducing inference overhead while preserving the strong reasoning abilities imparted by the Quiet-STaR framework.
    \item We introduce a curriculum learning-based multi-stage training strategy that progressively guides the model to learn a more concise thought trace, enabling it to internalize efficient reasoning patterns and express them compactly without performance degradation. We further accelerate Fast Quiet-STaR to the standard NTP-level setting via reinforcement learning-based fine-tuning, enabling implicit reasoning without explicit thought token generation.
    \item Extensive experiments show that Fast Quiet-STaR achieves comparable or even better performance than standard Quiet-STaR while reducing thought tokens. Fast Quiet-STaR NTP significantly outperforms the pre-trained model without increasing the inference time.
\end{itemize}

\section{Related Works}
\subsection{LLM Reasoning}
In recent years, enhancing the reasoning capabilities of large language models has become a major research focus \cite{rajani2019explain,zhang2025entropy,pan2025specreason}. The Chain-of-Thought (CoT) \cite{cot} prompting technique explicitly guides models to generate intermediate reasoning steps. Tree of Thoughts (ToT) \cite{tot} explores multiple reasoning paths through a tree-structured search. The CPO \cite{cpo} method combines ToT with Direct Preference Optimization (DPO) \cite{dpo}, using reasoning paths generated by ToT as paired training data to directly optimize the model’s CoT abilities. Self-Consistency \cite{wang2022self} samples multiple reasoning paths for the same problem and selects the final answer through a voting mechanism. Methods based on Monte Carlo Tree Search (MCTS) \cite{mcts} introduce classical planning algorithms into the reasoning process. Coconut \cite{hao2024training} explores the potential of unconstrained reasoning in latent spaces, highlighting the structural thinking capabilities of LLMs. 

In the latest research, reinforcement learning (RL) \cite{ppo} has emerged as a new paradigm for enhancing LLM reasoning. OpenAI's o1 \cite{openaio1} achieves significant improvements in reasoning performance. Similarly, models such as DeepSeek-R1 \cite{deepseekr1}, Kimi 1.5 \cite{Kimi1.5}, and QWQ \cite{QWQ} incorporate reinforcement learning \cite{shao2024deepseekmath} into pretrained models, exhibiting strong reasoning abilities.

Unlike most approaches that prompt models to ``think before answering'' on a per-question basis, Quiet-STaR \cite{quiet-star} shifts the reasoning process to a finer-grained, token-level paradigm. By encouraging deep reasoning at every token generation step, Quiet-STaR further enhances reasoning quality. However, as it requires long-range reasoning at every token, it incurs substantial inference latency, which limits its applicability in real-world scenarios. 
\subsection{Curriculum Learning}
Curriculum Learning is a training strategy that organizes the learning process by first presenting simpler examples and gradually introducing more complex ones. In recent years, curriculum learning has been widely adopted in the training of Large Language Models (LLMs) \cite{xu2020curriculum,nair2024curriculum}. LDCAL \cite{li2024active} leverages LLMs themselves to assess the difficulty of training instances, guiding the model to learn in an easy-to-hard sequence. TAPIR \cite{yue2024distilling} constructs a task-aware curriculum scheduling framework that dynamically adjusts the task distribution and progressively increases task complexity. Moreover, curriculum learning has also been employed to improve LLMs' understanding of long contexts by gradually increasing the context window size during training \cite{grattafiori2024llama}. Kimi 1.5 \cite{Kimi1.5} integrates a curriculum learning strategy during the reinforcement learning stage, allowing the model to start with simpler question before transitioning to more complex ones.

Unlike existing studies that apply curriculum learning at the data or task scheduling level, our approach integrates curriculum learning into the token-level reasoning process. By combining this strategy with the Quiet-STaR inference paradigm, our Fast Quiet-STaR better learns reasoning behaviors under limited Thought Tokens. 

\begin{figure*}[t]
  \centering
  \includegraphics[width=0.9\linewidth]{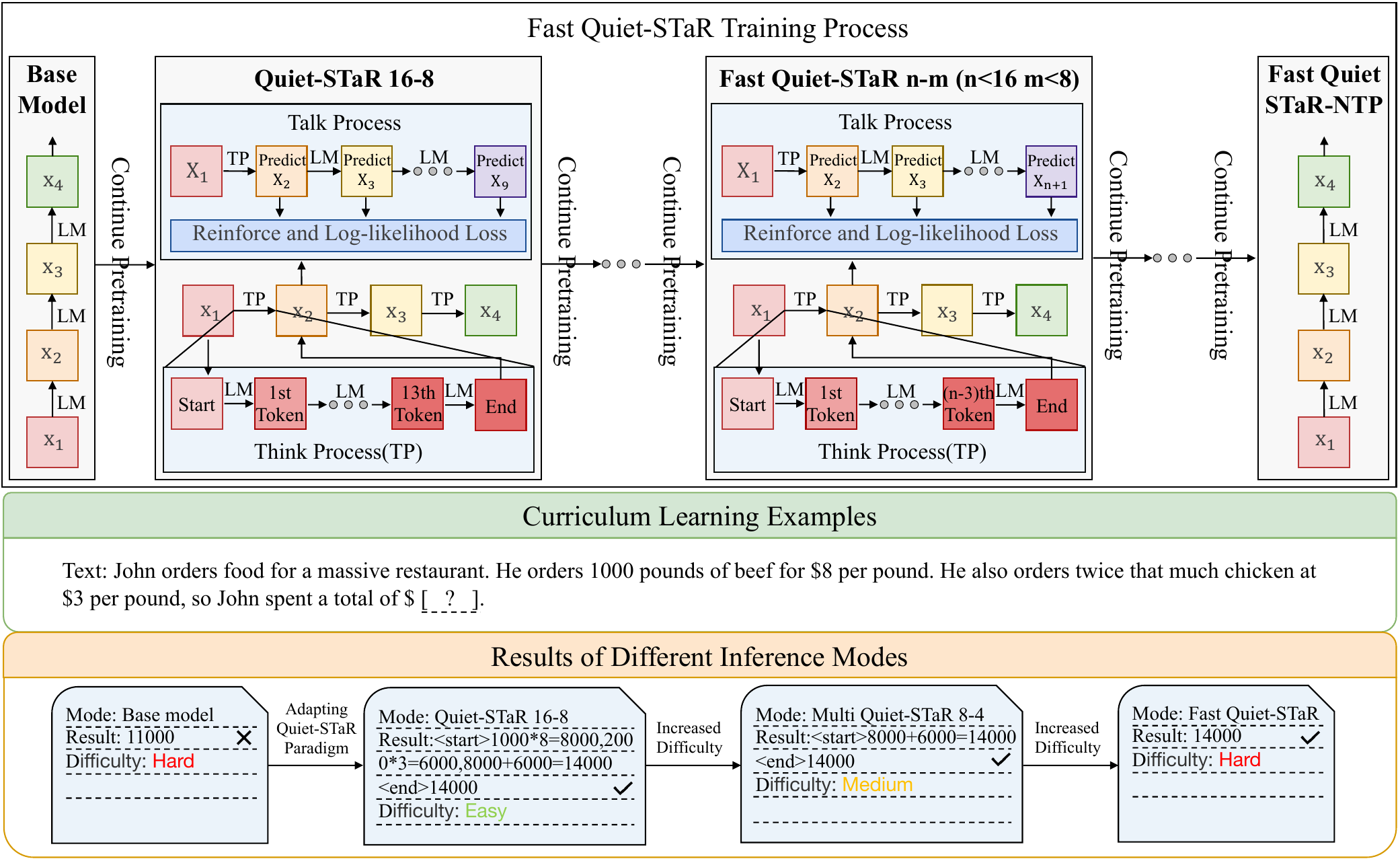}
  \caption{Fast Quiet-STaR training pipeline and Curriculum Learning Examples.} 
  \label{image1}
\end{figure*}
\section{Methodology}
The training procedure of Fast Quiet-STaR is illustrated in Figure \ref{image1}. Building upon Quiet-STaR, we propose a progressive, multi-stage training framework inspired by the principles of curriculum learning. This approach facilitates a gradual transition from easy to hard reasoning paradigms. In particular, during the final stage of training, we incorporate reinforcement learning to transition the reasoning paradigm of Quiet-STaR to the the standard NTP paradigm. 
\subsection{Quiet-STaR}
Quiet-STaR \cite{quiet-star} is a method for enabling language models to autonomously learn to generate internal rationales—referred to as ``thoughts''—in order to improve their ability to predict future tokens.The training process consisting of three distinct phases—Think, Talk, and Learn. 
\subsubsection{Think Process}
Given a token sequence $X=\{x_0,x_1,...,x_t\}$, Quiet-STaR \textbf{n-m} (n and m represent the number of thought/ahead tokens, respectively) generates a corresponding thought of length \textbf{n-1}, i.e., $T_i=(t_{i1},t_{i2},...,t_{i(n-1)})$, after each token $x_i$. Each thought is enclosed by learned meta-tokens \textit{<|start\_of\_thought|>} and \textit{<|end\_of\_thought|>}, which serve to activate and terminate the generation of thought, respectively. This process is executed in parallel using a custom attention mask, ensuring that each generated thought attends only to the corresponding prefix of the input sequence and the previously generated tokens within the same thought, denoted as: $P(t_{ij}|x_1,...,x_{i-1},\textit{<|start\_of\_thought|>},...,t_{i(j-1)})$. Note that m will be introduced in the Learn Process section. 
\subsubsection{Talk Process}
Quiet-STaR introduces a learnable interpolation mechanism. By employing a shallow MLP head to compute an interpolation weight $w$, conditioned on the hidden states of both the \textit{<|end\_of\_thought|>} token and the original input tokens. This weight modulates the influence of post-thought logits on the final prediction. The resulting mixed log-probability is defined in Equation~\ref{eq:one}.
\begin{equation}
  \label{eq:one}
  \log p_{i}^{\text {talk }}=w_{i} \log p_{i}^{\text {base }}+\left(1-w_{i}\right) \log p_{i}^{\text {thought}}
\end{equation}
Among them, $p_{i}^{\text{base}}$ represents the logits before thought, and $p_{i}^{\text {thought}}$ represents the logits after thought, w.r.t the token $x_i$. 
\subsubsection{Learn Process}
Quiet-STaR leverages the REINFORCE algorithm \cite{phan2023training} to optimize thought selection based on utility. It maximizes the log-likelihood of the next \textbf{m} (the number of ahead tokens) ground-truth tokens $X_{j+1: j+m+1}$ given prior context and a candidate rationale $T_j$. To reduce variance, multiple rationale continuations are sampled per token. The reward $r_j$ for each $T_j$ is defined as the difference between its log-likelihood $\log p_{j: j+m}^{\text{talk}}$ and the mean log-likelihood across all sampled rationales (see Eq.~\ref{eq:two}). Quiet-STaR incorporates this reward
\begin{multline}
  \label{eq:two}
  r_{j}=\log p_{j: j+m}^{\text {talk }}\left(X_{j+1: j+m+1}\right) \\
  -\log \bar{p}_{j: j+m}^{\text {talk }}\left(X_{j+1: j+m+1}\right)
\end{multline}
into a REINFORCE loss to update the model parameters $\theta$, encouraging thoughts that exceed the average, as shown in Equation~\ref{eq:three}. Additionally, Quiet-STaR includes a log-likelihood loss term, denoted as $L_{i}^{NLL}$, to ensure that the model not only learns to optimize the talking-head but also continues to receive next-token prediction signals for the base language model head.
\begin{multline}
  \label{eq:three}
  \nabla_{\theta} \mathcal{L}_{j}^{\text {REINFORCE }}= \\
  -r_{j} \cdot \nabla_{\theta} \log p_{\theta}\left(T_{j} \mid\left[X_{: j} ;\textit{<|start\_of\_thought|>}\right]\right)
\end{multline}
\subsection{Fast Quiet-STaR}
\subsubsection{Fast Quiet-STaR}
Compared with the mainstream NTP reasoning paradigm, Quiet-STaR introduces a new reasoning mechanism of "think first, talk later" for each token. Since it allows thinking, this mechanism effectively reduces the difficulty of predicting the next token, making the model perform better. Within the Quiet-STaR framework, a key hyperparameter, $n$, denotes the number of thought tokens, which has a significant impact on model performance. As illustrated in Figure \ref{Intro}, the model exhibits stronger reasoning abilities when more thought tokens are provided, while its performance degrades noticeably as the number of thought tokens decreases. 

This phenomenon raises a critical question: \textbf{why does a reduced number of thought tokens significantly impair the performance of Quiet-STaR, and how can we minimize the use of thought tokens without compromising model performance?} To investigate this, we analyze the thought traces under varying numbers of thought tokens. As shown in the lower part of Figure \ref{image1}, reducing the number of thought tokens forces the model to complete the reasoning process within a shorter sequence. This poses greater demands on the model’s ability to compress its reasoning steps, presenting a more challenging inference setting. In contrast to learning under easier setting (with more thought tokens), directly training the model on more difficult ones (with fewer thought tokens) proves less effective. This observation aligns with a core insight from curriculum learning: models often struggle to learn effectively when exposed to high-difficulty tasks early in training. Therefore, adopting a curriculum learning strategy, which progressively trains Quiet-STaR from easier settings to harder ones, holds promise for reasoning performance with limited thought tokens.

Based on the above observations, we adopt a curriculum learning strategy to facilitate the acquisition of reasoning paradigms and propose Fast Quiet-STaR approach. This method decomposes the training process into multiple stages, each aligned with a specific level of reasoning difficulty and corresponding modeling objective.

In the initial stage, the model is trained with a larger number of thought tokens. This setting represents a easy reasoning setting. As training progresses, we gradually reduce the number of thought tokens, thereby encouraging the model to engage in more concise and abstract reasoning under increasingly constrained resources. This encourages the model to progressively adapt to more difficult reasoning setting, enhancing both its reasoning efficiency and its generalization capabilities in lower-resource scenarios. Specifically, we begin with 16 thought tokens and 8 ahead tokens (16-8). During training, we gradually reduce it to 12-4 and 8-4.

\subsubsection{Fast Quiet-STaR NTP}
Although the number of thought tokens has been reduced, inference based on the Quiet-STaR paradigm still requires significantly higher computational resources compared to the NTP approach. To address this issue, we adopt reinforcement learning \cite{phan2023training} to transition the model's inference paradigm from Fast Quiet-STaR to NTP, called Fast Quiet-STaR NTP. Specifically, we initialize an NTP model using the checkpoint obtained from the last stage of the multi-stage training that includes 8 thought tokens and 4 ahead tokens. The log-likelihood loss \textbf{after thinking} of this checkpoint serves as a reference for computing rewards in reinforcement learning. The process of calculating reward is as follows: 
\begin{equation}
  \label{eq:five}
  r_j=\mathcal{L}_{FastQuietSTaR}-\mathcal{L}_{FastQuietSTaR-NTP}
\end{equation}
where $\mathcal{L}_{FastQuietSTaR}$ represents the negative log-likelihood loss of Fast Quiet-STaR 8-4 at the $j$th token after a thinking process, and $\mathcal{L}_{FastQuietSTaR-NTP}$ represents the negative log-likelihood loss of Fast Quiet-STaR NTP at the $j$th token.  The final loss function is as follows:
\begin{equation}
  \label{eq:six}
  \nabla_{\theta} \mathcal{L}_{j}^{\text {REINFORCE }}= -r_{j} \cdot \nabla_{\theta} \log p_{\theta}\left(x_{j} \mid X_{: j} \right)
\end{equation}
Through reinforcement learning, the model is encouraged to emulate the prediction quality of Fast Quiet-STaR model without explicitly generating intermediate reasoning tokens during inference. Notably, this transition enables Fast Quiet-STaR NTP to effectively internalize the reasoning process, compressing and integrating the previously explicit ``thinking'' into its latent representations.

\section{Experiments}
\begin{table*}[t]
    \centering
    \caption{Performance (\%) comparison. \textbf{Bold} and \underline{underline} denote the best and second-best performance of models. For each method, we report their time to first token (TTFT, in seconds). Performance \textbf{$\Delta$} represents the difference between Fast Quiet-STaR NTP and Pre-Trained.}
    \vskip 0.15in
    \label{tab:main}
    \resizebox{0.95\textwidth}{!}{\begin{tabular}{c|ccc|cccc|c}
\toprule
Method & Thought Tokens & Ahead Tokens & TTFT (s) & PIQA & SIQA & CommonsenseQA & GSM8K & AVG \\ 
\midrule
\multicolumn{8}{c}{\textbf{Mistral-7B}}  \\ 
\midrule
Pre-Trained & 1 & 1  & \textbf{0.028}   & 45.9  & 41.6 & 35.4   & 4.9 & 32.0 \\
Quiet-STaR & 16 & 8  & 0.738   & 54.7  & 47.0 & 45.3   & \underline{9.8} & 39.2 \\ \hline
Quiet-STaR & 12 & 4  & 0.550  & 53.1  & 45.7   & 43.4 & 8.4 & 37.7 \\ 
Fast Quiet-STaR & 12 & 4  & 0.550  & \textbf{59.0} & \textbf{52.5}  & \textbf{50.7}   & \textbf{10.0} & \textbf{43.1} \\ \hline
Quiet-STaR & 8 & 4  & \underline{0.305}  & 49.1  & 42.2 & 39.3   & 7.2 & 34.5 \\ 
Fast Quiet-STaR & 8 & 4  & \underline{0.305}  & \underline{56.9} & \underline{51.1} & 49.0   & \underline{9.8} & \underline{41.7} \\ \hline
Fast Quiet-STaR-NTP & 1 & 1  & \textbf{0.028} &  55.0  & 50.1   & \underline{49.2} & 9.6 & 41.0 \\ 
\rowcolor[RGB]{220, 240, 255} Performance \textbf{$\Delta$}  & - & -  & - &  \textbf{{\color{red} +9.1}}  & \textbf{{\color{red} +8.5}}  & \textbf{{\color{red} +13.8}} & \textbf{{\color{red} +4.7}} &  \textbf{{\color{red} +9.0}} \\ 
\midrule
\multicolumn{8}{c}{\textbf{Qwen2.5-7B}}  \\ 
\midrule
Pre-Trained & 1 & 1  & \textbf{0.026} & 70.1  & 60.7 & 52.4   & 11.6 & 48.7 \\
Quiet-STaR & 16 & 8  & 0.633 & \textbf{77.6}  & \textbf{68.1} &  \textbf{66.5}  & \textbf{17.7} &  \textbf{57.5} \\ \hline
Quiet-STaR & 12 & 4  & 0.481  & 72.4  & 61.8 &   59.7 & 17.1 & 52.8 \\ 
Fast Quiet-STaR & 12 & 4  & 0.481  & 74.3 & 64.5  &  \underline{63.9}  & \underline{17.6} & \underline{55.1} \\ \hline
Quiet-STaR & 8 & 4  & \underline{0.269}  & 70.2  & 60.3 &  54.9  & 11.9 & 49.3 \\ 
Fast Quiet-STaR & 8 & 4  & \underline{0.269} & 74.5 & 63.4 &  59.3 & 16.9 & 53.5 \\ \hline
Fast Quiet-STaR-NTP & 1 & 1  & \textbf{0.026} & \underline{74.9} & \underline{65.8}  & 60.3  & 16.5 & 54.4 \\ 
\rowcolor[RGB]{220, 240, 255} Performance \textbf{$\Delta$} & - & -  & - & \textbf{{\color{red} +4.8}} & \textbf{{\color{red} +5.1}}  & \textbf{{\color{red} +7.9}}  & \textbf{{\color{red} +4.9}} & \textbf{{\color{red} +5.7}} \\ 

\bottomrule
\end{tabular}}
\vskip -0.1in
\end{table*}
\begin{table}[t]
    \centering
    \caption{Comparison of generation latency(in seconds) between different methods. For prefix length 256 and generate length 128, we use a prompt of 256 tokens and let the model generate 128 tokens after the prompt. AVG ACC represents the average accuracy on PIQA, SIQA, CommonsenseQA and GSM8K. } 
    \vskip 0.15in
    \label{tab:time}
    \resizebox{0.45\textwidth}{!}{\begin{tabular}{c|cc|cc|c}
\toprule
Prefix Length & \multicolumn{2}{c|}{256} & \multicolumn{2}{c|}{512} & \multicolumn{1}{c}{AVG}   \\ 
Generate Length & 128  & 256  & 256 & 512 &  ACC   \\
\midrule
Pre-Trained & \textbf{3.2}   & \textbf{7.3}  & \textbf{8.8}   & \textbf{17.1} & \textbf{32.0}  \\
Quiet-STaR 16-8 &   52.7   & 116.9  & 167.0   & 326.4 & 39.2   \\ \hline
Quiet-STaR 12-4 &   40.6   & 92.9  & 102.4   & 288.6 & 37.7     \\ 
Fast Quiet-STaR 12-4 &   40.6   & 92.9  & 102.4   & 288.6 & 43.1   \\ 
\hline
Quiet-STaR 8-4 &  33.0   & 65.9  & 82.4   & 184.4 & 34.5  \\ 
Fast Quiet-STaR 8-4 &  33.0   & 65.9  & 82.4   & 184.4 & 41.7   \\ 
\hline
\textbf{Fast Quiet-STaR-NTP} & \textbf{3.2}   & \textbf{7.3}  & \textbf{8.8}   & \textbf{17.1} & \textbf{41.0} \\ 
\bottomrule
\end{tabular}}
\vskip -0.1in
\end{table}
\subsection{Experimental Settings}
\textbf{Post-Training Settings.} We perform post-training on Mistral 7B \cite{mistral7b} and Qwen2.5 7B \cite{qwen25} using the OpenWebMath dataset \cite{paster2023openwebmath} and evaluate its ability to directly predict answers on the CommonsenseQA \cite{talmor2018commonsenseqa} and GSM8K \cite{gsm8k} benchmarks. Following \cite{quiet-star}, we calculate the accuracy rate as: $ACC=\frac{\prod_{i=0}^{l}P(A_i|Q_1,...,Q_k,A_i,...A_{i-1})}{\prod_{i=0}^{l}(\sum_{A_j\in S_{ans}}(P(A_j|Q_1,...,Q_k,A_i,...A_{i-1})) )}$, where $Q_i$ represents the question token, $A_i$ represents the answer token, $k$ and $l$ represent their lengths, and $S_{ans}$ represents the candidate set of answers (e.g. $S_{ans}=\{A,B,C,D,E\}$ for CommonsenseQA). This evaluation and training setup is consistent with the Quiet-STaR \cite{quiet-star}. To further assess the effectiveness of Fast Quiet-STaR in general reasoning tasks, we also introduce two more general-purpose evaluation benchmarks: SIQA \cite{siqa} and PIQA \cite{piqa}. 

\textbf{Implementation details.} All training experiments are conducted on 8 H800 GPUs. For Quiet-STaR, we train for 100 steps, and for Fast Quiet-STaR, we select the last checkpoint in the previous stage for initialization and train for another 50 steps for each training stage. See Appendix \ref{sec:Implementation details} for more details. 
\subsection{Main Results}
We evaluate Quiet-STaR, Fast Quiet-STaR, and Fast Quiet-STaR NTP on four benchmarks: PIQA \cite{piqa}, SIQA \cite{siqa}, CommonsenseQA \cite{talmor2018commonsenseqa}, and GSM8K \cite{gsm8k} (Table~\ref{tab:main}). Under \textbf{equal TTFT}, Fast Quiet-STaR consistently outperforms Quiet-STaR, exhibiting stable performance even as the number of thought tokens decreases—unlike Quiet-STaR, which degrades significantly. For Mistral 7B, multi-stage training further boosts performance: Fast Quiet-STaR with 8 tokens surpasses the 16-token variant by 1.8\% while cutting inference time to 41.3\%. Compared to pre-trained baselines, Fast Quiet-STaR NTP achieves notable gains without added compute, improving average accuracy by 9\% on Mistral 7B and 5.7\% on Qwen2.5 7B. These results validate the effectiveness of incorporating a curriculum learning strategy within the Quiet-STaR framework to simultaneously improve both model efficiency and performance.

We further analyze generation latency, a more general metric for evaluating speed. For prefix lengths of 256 and 512, the models generate 128/256 and 256/512 tokens, respectively (Table~\ref{tab:time}). Fast Quiet-STaR NTP significantly reduces latency, achieving just 6\% of the end-to-end generation time of Quiet-STaR 16-8 (for 256-128), on par with the pre-trained baseline. Additionally, it yields accuracy gains of 0.8\% and 9\%, respectively. These results highlight Fast Quiet-STaR NTP effectiveness in addressing both the latency of Quiet-STaR and the poor performance of standard pre-trained models.
\subsection{Experimental Analysis}
We choose Mistral 7B \cite{mistral7b} for our analytical experiments, which is consistent with Quiet-STaR\cite{quiet-star}. 
\subsubsection{Ablation Studies}
\begin{figure}[t]
  \centering
  \includegraphics[width=1\linewidth]{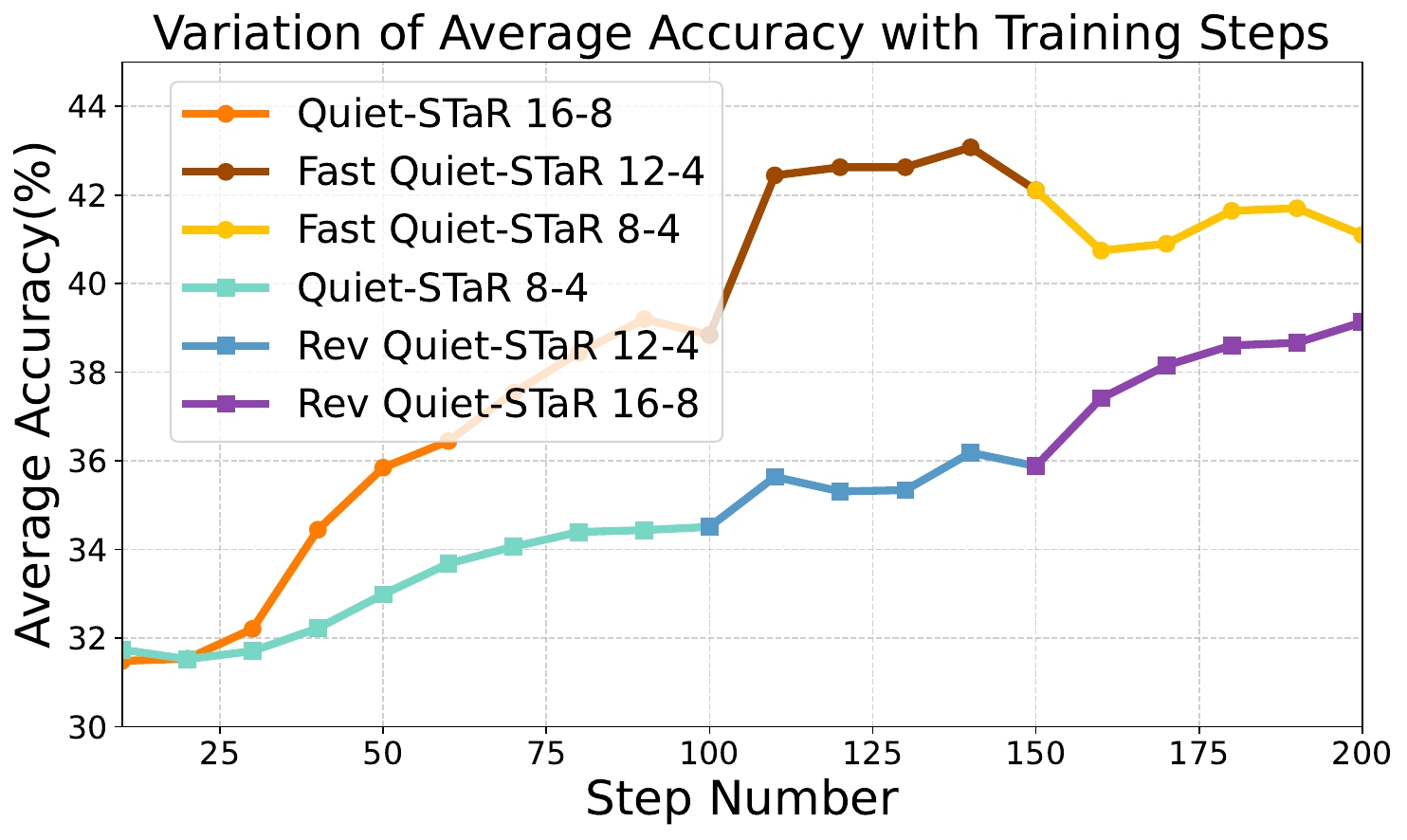}
  \caption{Comparison of the average accuracy of Fast Quiet-STaR and Rev Quiet-STaR with training steps.} 
  \label{image2}
\end{figure}

\textbf{Curriculum Learning.} To evaluate the effectiveness of our easy-to-hard multi-stage curriculum learning training strategy, we experiment with an alternative where we reverse the entire training process. Specifically, we start from the Quiet-STaR 8-4 model and follow a ``8-4 → 12-4 → 16-8'' training sequence. At each stage, the model is initialized with the weights obtained from the previous stage. We refer to this series of progressively trained models as Rev Quiet-STaR. We compare the average performance of Rev Quiet-STaR and Fast Quiet-STaR across four benchmarks: PIQA, SIQA, CommonsenseQA, and GSM8K (Figure~\ref{image2}).  Experimental results indicate that the multi-stage training method, progressing from difficult to easier, does not lead to performance improvements. Notably, Rev Quiet-STaR 16-8 even underperforms Fast Quiet-STaR 8-4, despite utilizing a larger number of thought tokens. 

\begin{figure}[t]
  \centering
  \includegraphics[width=1\linewidth]{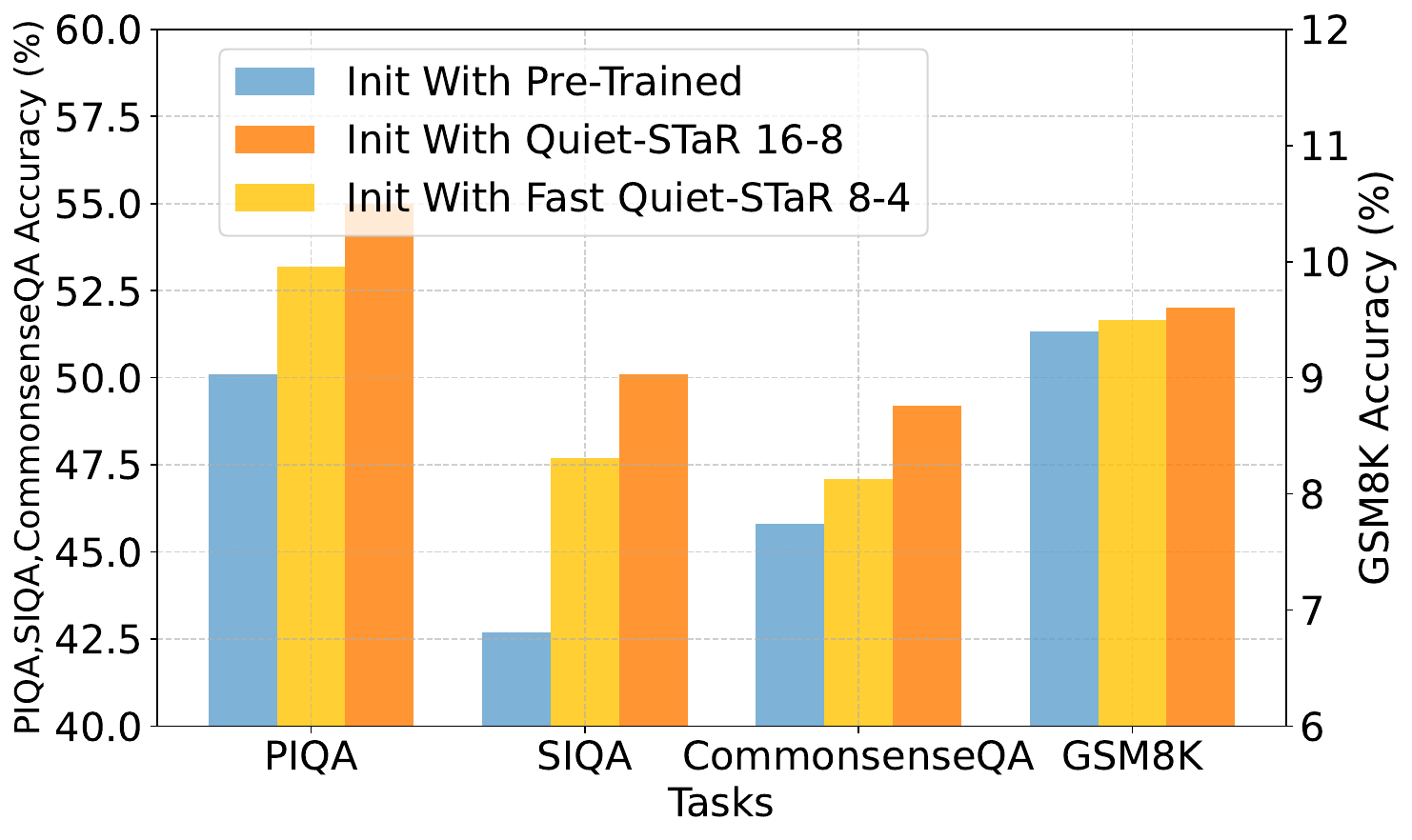}
  \caption{Comparison of Fast Quiet-STaR NTP under different initialization schemes.The left vertical axis corresponds to the average accuracy on PIQA, SIQA, and CommonsenseQA, while the right vertical axis indicates the accuracy on GSM8K.}
  \label{image3}
\end{figure}

\textbf{Reinforcement Learning Initialization.} To study the impact of initialization, we compare Fast Quiet-STaR 8-4 with two alternatives: the pre-trained model and Quiet-STaR 16-8. We evaluate all approaches on four benchmarks—PIQA, SIQA, CommonsenseQA, and GSM8K—summarized in Figure~\ref{image3}. Results show that Fast Quiet-STaR 8-4 yields the best performance, followed by Quiet-STaR 16-8, and then the pre-trained model. We attribute this to Fast Quiet-STaR 8-4’s ability to generate a compact yet informative thought trace, which is conducive to further improving efficiency and expanding the reasoning paradigm to NTP. In contrast, the pre-trained model lacks an explicit thought trace prior; Quiet-STaR 16-8 provides detailed thought traces, which rely on a longer reasoning process, which may lead to a larger span of reasoning paradigm difficulty when learning, thus affecting the overall training performance.
\begin{table}[t]
    \centering
    \caption{Performance comparison between Quiet-STaR NTP and Fast Quiet-STaR NTP. CSQA represents for CommonsenseQA, and Performance \textbf{$\Delta$} represents the difference between Fast Quiet-STaR NTP and Quiet-STaR NTP.  }
    \vskip 0.15in
    \label{tab:ntp}
    \resizebox{0.45\textwidth}{!}{\begin{tabular}{c|ccccc}
\toprule
Method & PIQA & SIQA & CSQA & GSM8K & AVG \\ 
\midrule
Quiet-STaR NTP & 49.1 & 44.3  & 42.5   & 7.3  & 38.1  \\
Fast Quiet-STaR NTP & 55.0  & 50.1 &  49.2  & 9.6 & 41.0 \\ 
\rowcolor[RGB]{220, 240, 255} Performance \textbf{$\Delta$} & \textbf{{\color{red} +5.9}}  & \textbf{{\color{red} +5.8}} &  \textbf{{\color{red} +6.7}}  & \textbf{{\color{red} +2.3}} & \textbf{{\color{red} +5.2}} \\ 
\bottomrule
\end{tabular}}
\vskip -0.1in
\end{table}

\textbf{Fast Quiet-STaR NTP Without Curriculum Learning.} To evaluate the effectiveness of the curriculum learning procedure ``16-8 → 12-4 → 8-4 → NTP'', we omit the intermediate stages. Specifically, we directly initialize the pre-trained model with Quiet-STaR 16-8 and use its log-likelihood loss as the reference for computing rewards in reinforcement learning, resulting in Quiet-STaR NTP. As shown in Table \ref{tab:ntp}, this shortcut results in a 5.2\% drop in average accuracy compared to Fast Quiet-STaR NTP obtained through the full curriculum. These results underscore the critical role of the curriculum learning process in enhancing the overall performance of the model. 
\begin{figure}[t]
  \centering
  \includegraphics[width=1\linewidth]{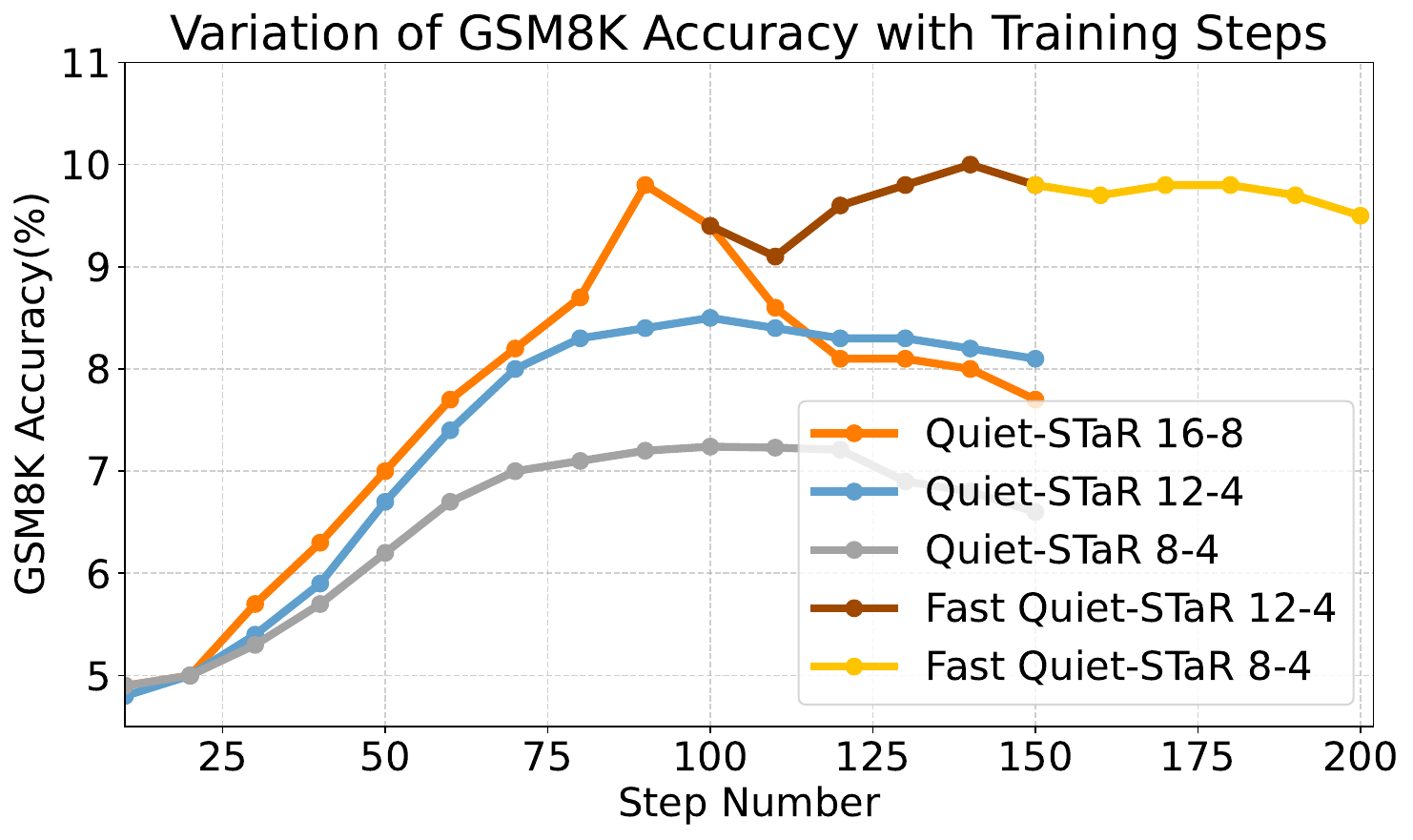}
  \caption{Comparison of the accuracy of Quiet-STaR and Fast Quiet-STaR on GSM8K during training.}
  \label{Step}
\end{figure}
\subsubsection{Data Efficiency} 
To ensure that Fast Quiet-STaR's performance gains do not stem from an increase in data volume, we track GSM8K accuracy throughout training (Figure~\ref{Step}). Quiet-STaR trains for 150 steps, with performance peaking around step 100 and declining thereafter—consistent with prior findings \cite{quiet-star}. In contrast, Fast Quiet-STaR achieves strong results with just 20–40 additional steps. These results suggest that the gains arise from the progressive learning mechanism of multi-stage training, not from greater data exposure. 
\begin{table}[t]
    \centering
    \caption{Zero-shot performance on Fast Quiet-STaR and Pre-Trained applied to chain-of-thought on GSM8K. Performance \textbf{$\Delta$} represents the difference between Fast Quiet-STaR NTP and Pre-Trained. }
    \vskip 0.15in
    \label{tab:gsm_cot}
    \resizebox{0.45\textwidth}{!}{\begin{tabular}{c|ccccc}
\toprule
Method & maj@2 & maj@3 & maj@4 & maj@5 & maj@6 \\ 
\midrule
Pre-Trained & 28.5 & 32.6  & 37.5   & 40.3  & 43.3  \\
Fast Quiet-STaR NTP & 36.0  & 40.6 &  45.8  & 49.2 & 52.4 \\ 
\rowcolor[RGB]{220, 240, 255} Performance \textbf{$\Delta$} & \textbf{{\color{red} +7.5}}  & \textbf{{\color{red} +8.0}} &  \textbf{{\color{red} +8.3}}  & \textbf{{\color{red} +8.9}} & \textbf{{\color{red} +9.1}} \\ 
\bottomrule
\end{tabular}}
\vskip -0.1in
\end{table}
\subsubsection{Performance on Generation Tasks} 
To evaluate the performance of Fast Quiet-STaR on generative tasks, we compare Fast Quiet-STaR NTP with the original pre-trained model under the Next Token Prediction (NTP) inference paradigm on the GSM8K dataset. Specifically, we adopt the Chain-of-Thought (CoT) reasoning approach and measure accuracy using majority voting over 6 samples (cot-maj@6), with results as show in table \ref{tab:gsm_cot}. Experimental results show that as the number of votes increases, the performance advantage of Fast Quiet-STaR NTP over the pretrained model becomes more pronounced. On the cot-maj@6 metric, Fast Quiet-STaR NTP achieves an accuracy improvement from 43.3\% to 52.4\%, demonstrating its effectiveness in complex reasoning tasks. These results demonstrate that Fast Quiet-STaR can further enhance inference performance on top of CoT reasoning and \textbf{Fast Quiet-STaR is complementary to CoT}, rather than redundant.
\subsubsection{Analysis of long-term inference savings}
The training process of Fast Quiet-STaR is highly efficient. The entire training pipeline (from Quiet-STaR 16-8 → Fast Quiet-STaR 12-4 → Fast Quiet-STaR 8-4 → Fast Quiet-STaR NTP) requires only 0.5M tokens for continue pre-training. This process takes just 54 minutes on 8 H800 GPUs.
We use prompts containing 256 tokens and allow the model to generate 128 tokens for demonstration and analysis purposes. On a single H800 GPU, the inference latency of Fast Quiet-STaR-NTP is 3.2 seconds, compared to 52.7 seconds for Quiet-STaR 16-8(more scenarios with inference latency are in Table \cite{tab:time}). For 67 end-to-end inference runs, Quiet-STaR 16-8 requires approximately 59 minutes in total, whereas Fast Quiet-STaR-NTP completes the same task in just 4 minutes—yielding an overall speedup of 55 minutes. In other words, the time saved from just 67 inference runs is sufficient to offset the entire training cost of Fast Quiet-STaR-NTP. More importantly, in real-world deployments, models are typically required to perform millions of inference runs—far exceeding 67—where our approach would demonstrate substantial advantages in total inference cost.
\subsubsection{Thought token analysis}
\label{Thought_ana}
\begin{figure}[t]
  \centering
  \includegraphics[width=1\linewidth]{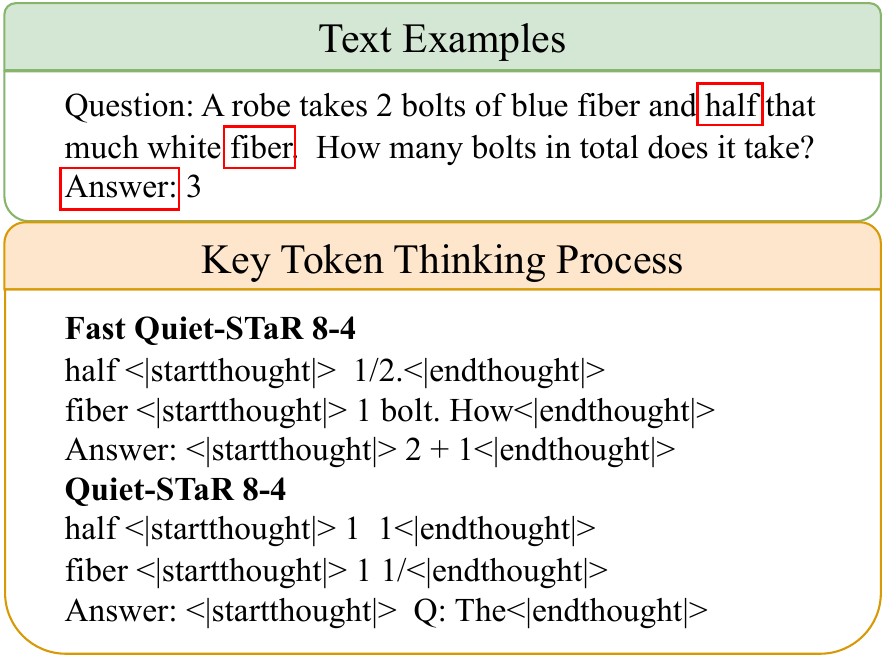}
  \caption{Examples of the text and its thought process at key tokens.}
  \label{token_ana}
\end{figure}

We visualize the thought tokens generated by Quiet-STaR 8-4 and Fast Quiet-STaR 8-4 at key positions—tokens most informative for final predictions—on the GSM8K dataset to examine their internal reasoning behavior. As shown in Figure~\ref{token_ana}, Quiet-STaR 8-4 produces relatively unstructured thoughts, indicating incomplete acquisition of the Quiet-STaR reasoning paradigm. In contrast, Fast Quiet-STaR 8-4 demonstrates more abstract and goal-directed reasoning behavior. These observations indicate that the incorporation of a curriculum learning strategy—progressing from easier to more difficult—enables Fast Quiet-STaR to gradually acquire the ability to perform effective reasoning under resource constraints. For more examples, please refer to Appendix \ref{sec:Thought}.
\section{Conclusion}
In this paper, we proposed Fast Quiet-STaR, an efficient extension of the Quiet-STaR reasoning paradigm that maintains the core benefits of fine-grained token-level reasoning while significantly reducing inference overhead. By leveraging a curriculum learning-based training strategy that progressively reduces the number of thought tokens, Fast Quiet-STaR enables models to develop compact yet effective reasoning abilities. Furthermore, through reinforcement learning-based fine-tuning, we extend this paradigm to the standard Next Token Prediction setting, eliminating the need for explicit thought-token generation during inference. Experiments on Mistral 7B and Qwen2.5 7B across four benchmark datasets show that Fast Quiet-STaR achieves substantial gains over Quiet-STaR under the same number of thought tokens, and Fast Quiet-STaR NTP outperforms the pre-trained model and performs on par with Quiet-STaR. These results highlight Fast Quiet-STaR as a practical solution for enhancing reasoning capabilities in LLMs.
\section*{Limitations}

Despite the notable improvements in inference efficiency and performance achieved by Fast Quiet-STaR, several limitations remain. First, the evaluation in this study primarily focuses on mathematical and logical reasoning tasks, leaving its generalization capability to other domains yet to be thoroughly validated. Second, this method is only for the Quiet-STaR reasoning method.

\bibliography{custom}

\appendix

\section{Implementation details}
\label{sec:Implementation details}
Our experimental settings are consistent with Quiet-STaR \cite{quiet-star}. Specifically, we employ the AdamW optimizer with a warm-up step count of 20, a weight decay of 0.001, and a batch size of 8. For Quiet-STaR, we train for 100 steps, and for Fast Quiet-STaR, we select the last checkpoint in the previous stage for initialization and train for 50 steps for each stage. The learning rates are adjusted slightly depending on the model: we use a learning rate of 1e-6 for Mistral \cite{mistral7b} and 8e-6 for Qwen2.5 \cite{qwen25}. During training, we perform sampling with a temperature of T=1. For evaluation, we adopt greedy decoding to ensure deterministic outputs. All training experiments are conducted on eight H800 GPUs. For measuring the Time to First Token (TTFT), we utilize a single H800 GPU and fix the context length to 256. TTFT is defined as the elapsed time between the moment the model receives the full input sequence and the generation of the first token. 

We report the version numbers of used packages in Table \ref{tab:package_version}.
\begin{figure}[t]
  \centering
  \includegraphics[width=1\linewidth]{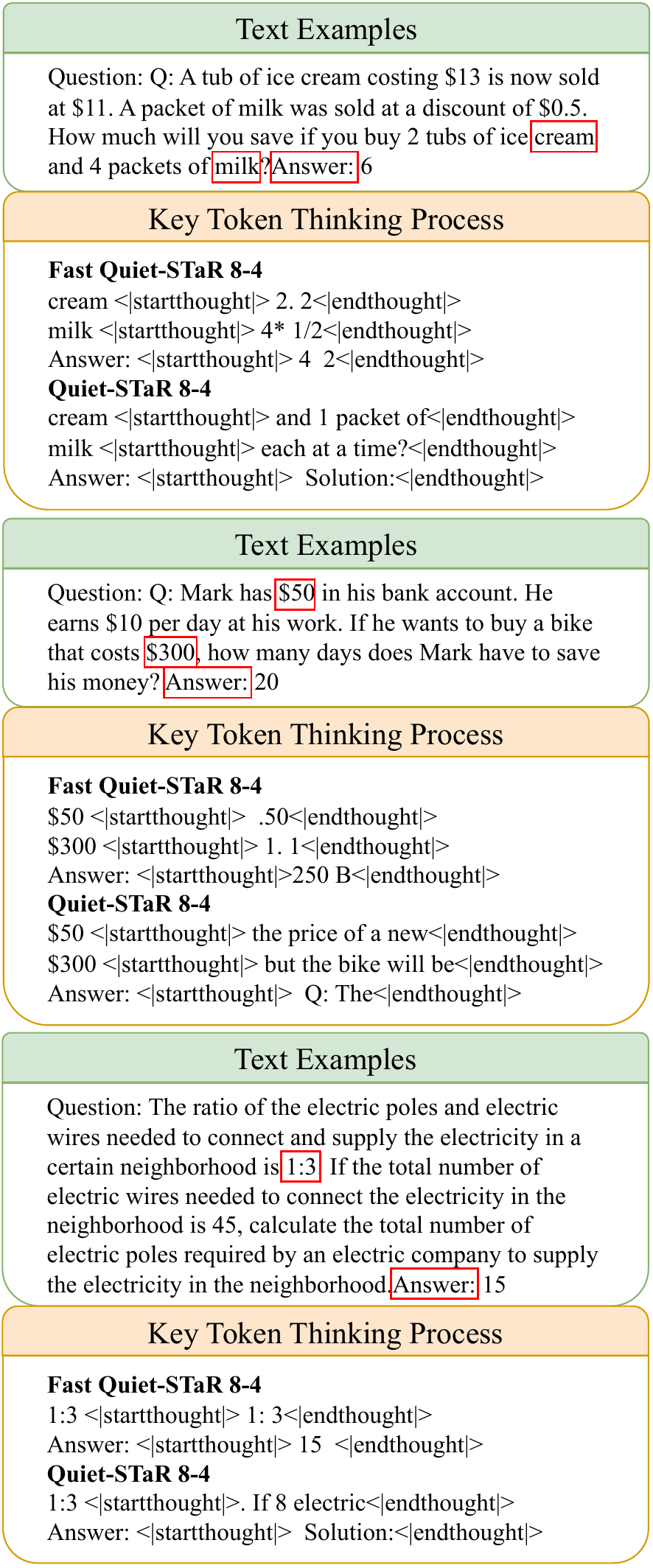}
  \caption{More examples of the text and its thought process at key positions.}
  \label{token_ana_more}
\end{figure}
\begin{table}[!h]
\centering
\scalebox{0.82}{\begin{tabular}
{lc|lc}
\toprule
 Package & Version & Package & Version \\
\midrule
PyTorch & 2.1.0 & transformers & 4.46.0 \\
deepspeed & 0.10.0 & tokenizers & 0.13.3 \\
datasets & 2.14.3 &  &  \\
\bottomrule
\end{tabular}}
\caption{Versions of used packages.}
\label{tab:package_version}
\end{table}
\section{More Thought token analysis}
\label{sec:Thought}
Based on the settings described in Section \ref{Thought_ana}, we further visualize examples of thought tokens generated at key positions by Quiet-STaR 8-4 and Fast Quiet-STaR 8-4, as illustrated in Figure \ref{token_ana_more}. Compared to Quiet-STaR 8-4, the thought tokens generated by Fast Quiet-STaR 8-4 exhibit a more abstract and directive reasoning process.
\section{License for Scientific Artifacts}
The Open web math \cite{paster2023openwebmath} is licensed under ODC-By 1.0 License\footnote{https://spdx.org/licenses/ODC-By-1.0.html}. 
The Mistral model \cite{mistral7b} and Qwen2.5 model \cite{qwen25} is licensed under Apache License 2.0 license\footnote{https://choosealicense.com/licenses/apache-2.0/}. The evaluation datasets \cite{gsm8k,talmor2018commonsenseqa,siqa,piqa} are subject to the MIT license\footnote{https://choosealicense.com/licenses/mit/}. All usages of scientific artifacts in this paper obey the corresponding licenses.
\end{document}